\documentclass[sigconf,authorversion,nonacm,natbib=false]{acmart}
\usepackage{graphicx} 
\graphicspath{ {./images/} }
\usepackage{array}
\usepackage{tabularx}
\usepackage{tikz}
\usetikzlibrary{shapes.geometric, arrows}
\tikzstyle{startstop} = [rectangle, rounded corners, minimum width=.25cm, minimum height=.25cm,text centered, draw=black, fill=red!30]
\tikzstyle{io} = [rectangle, minimum width=1cm, minimum height=.5cm, text centered, draw=black, fill=green!30]
\tikzstyle{process} = [rectangle, minimum width=1cm, minimum height=1cm, text centered, draw=black, fill=orange!30]
\tikzstyle{decision} = [rectangle, minimum width=.45cm, minimum height=1cm, text centered, draw=black, fill=blue!30]
\tikzstyle{arrow} = [thick,->,>=stealth]
\usepackage{varwidth}


\usepackage[
backend=biber,
style=alphabetic,
sorting=ynt,
]{biblatex} 
\usepackage{xcolor}         

\title{Sequencing Matters: A Generate-Retrieve-Generate Model for Building Conversational Agents}

\author{Quinn Patwardhan}\authornote{Work done during summer internship at Georgetown University.}
 \email{qroshan5@gmail.com}
 \affiliation{%
 \department{Sidwell Friends School}
  \institution{Washington, D.C., USA}
  \country{USA}
  }

  \author{Grace Hui Yang}
 \email{grace.yang@georgetown.edu}
 \affiliation{%
 \department{InfoSense, Dept. of Computer Science}
  \institution{Georgetown University, Washington, D.C., USA}
  \country{USA}
  }

\begin{document}

\maketitle


\section*{Abstract}
The Text Retrieval Conference (TREC)'s Interactive Knowledge Assistance (iKAT) Track has the goal of combining conversational and personalizable elements with existing information retrieval (IR) technologies to facilitate information-seeking. To accomplish this, an iKAT system is given two pieces of information from the user: 1) a Personal Textual Knowledge Base (PTKB), which is a persistent set of a handful of factual statements about the user (like "I am lactose intolerant" or "I am afraid of roller coasters") that lasts throughout a conversation, and 2) the user utterance, which is usually written from an information-seeking standpoint. In an automatic run, the system must find both the PTKBs relevant to each utterance and provide relevant responses to both the current utterance and the conversation history. Answers must be generated based on passages retrieved from the ClueWeb 22B Corpus. 

This paper contains what the Georgetown InfoSense group has done in regard to solving the challenges presented by TREC iKAT 2023. Our submitted runs outperform the median runs by a significant margin, exhibiting superior performance in nDCG across various cut numbers and in overall success rate. Our approach uses a Generate-Retrieve-Generate method, which we've found to greatly outpace Retrieve-Then-Generate approaches for the purposes of iKAT. Our solution involves the use of Large Language Models (LLMs) for initial answers, answer grounding by BM25, passage quality filtering by logistic regression, and answer generation by LLMs again. We leverage several purpose-built Language Models, including BERT, Chat-based, and text-to-transfer-based models, for text understanding, classification, generation, and summarization. The official results of the TREC evaluation contradict our initial self-evaluation, which may suggest that a decrease in the reliance on our retrieval and classification methods is better. Nonetheless, our findings suggest that the sequence of involving these different components matters, where we see an essentiality of using LLMs before using search engines.


\section{Introduction}

The Text Retrieval Conference (TREC) Interactive Knowledge Assistance (iKAT) Track has the goal of combining conversational and personalizable elements with existing information retrieval (IR) technologies to facilitate information-seeking. The iKat Track builds upon and replaces a previous track of TREC, CAST, whose goals were similar in that it also involved elements of both conversation and information retrieval. The new iKAT Track differs in that it adds the challenge of personalizing responses to the user, based on the PTKBs. A Personal Textual Knowledge Base (PTKB) is a persistent set of a handful of factual statements about the user (like "I am lactose intolerant" or "I am afraid of roller coasters") that lasts throughout a conversation. The iKAT Track is in the domain of both conversational agents and Task-based information seeking. 



The most common traditional approach to the iKAT challenge could be fully retrieval-based \cite{Retrieval_Models_for_Question_and_Answer_Archives}, where retrieval models are used to retrieve relevant passages and answers are extracted, summarized, or generated from these relevant passages. 
After the introduction of Large Language Models (LLMs) such as ChatGPT, many began using one Large Language Model to do everything, including using them for information retrieval and information seeking. This came with drawbacks, due to the issues LLMs face from hallucinations \cite{zhang2023sirens} and bias \cite{ferrara2023chatgpt}. 
Without grounding or correction, Large Language Models tend to hallucinate and state false information. Thus, their ability to generate factual responses, as required by iKAT, is limited. On the contrary, most retrieval methods work much better for finding relevant passages, although they lack the conversationality of LLMs. For the task of generating responses directly from the corpus, many rely on Extractive QA Models \cite{seonwoo2020contextaware}, although this method lacks some of the inference abilities of LLMs to combine their knowledge base with provided information to form a fluent response directly to the user utterance.

To augment the text generation capabilities of Large Language Models, researchers have started to work on combining LLMs' strengths with retrieval methods, so that LLMs' outputs can be backed up with more trustworthy and accurate content and grounded in corpora that they exist. These LLM Grounding methods can be grouped into three categories: {\it Fact-Checking}, {\it Retrieve-Then-Generate}, and {\it Generate-Then-Retrieve}. 

The {\it Fact-Checking} type works by fact-checking and citing the claims made by the LLM, and rewriting any claim identified as false and/or unciteable \cite{gao2023rarr}. In these approaches, a majority of the text content was LLM generated, which allows for the conversational and deeper language-understanding benefits of Language Models. However, this still leaves room for model bias, and the process of fact-checking and citing is not 100\% effective. Fact-checking also poses the problem of layered hallucinations, in which the LLM hallucinations both in the output text and when finding the citation, making incorrect information appear to be truthful and cited \cite{zuccon2023chatgpt}.  


The {\it Retrieve-Then-Generate} approaches \cite{ma2023pretraining} involves finding supporting information, then using an LLM to generate a response based on that information. We initially considered using a similar approach, although ran into issues with the LLM not solely drawing from given information and instead drawing mostly from its training data. This method does not guarantee full accuracy, and still has the same problems, although to a lesser extent, as just pure generation techniques. 

The {\it Generate-Then-Retrieve} methods \cite{huo2023retrieving} use a prompting technique, where they find supporting passages (either from using an LLM response or just the user utterance), then ask the LLM to answer the question from the passage. 


Our approach is mainly a {\it Generate-Retrieve-Generate} approach. We use the LLM to generate a response to the user utterance with minimal prompting (only providing the relevant PTKBs and utterance combined in a linguistically fluent manner). We encourage the LLM to generate a lengthy response, in order to get more information that can be used to find grounding materials. Even if the LLM states something incorrect, there is a high likelihood that its response will correspond significantly with what the user asked it. Once we've retrieved relevant passages, we then use a text-to-text transfer Large Language Model specifically designed for summarization to summarize each passage, after optimizing the sentences of each passage to prioritize the information actually relevant to the user utterance. Our method maintains the citability of fact-checking-based methods and the conversational fluency enabled by LLMs. Unlike the generate-then-retrieve methods, our approach still involved using an LLM in creating our final response, which allowed for more intelligent and fluent responses. This approach has been used to solve similar problems before, and has scored much higher on QA tests than other methods \cite{yu2023generate}.

\section{Proposed Method}


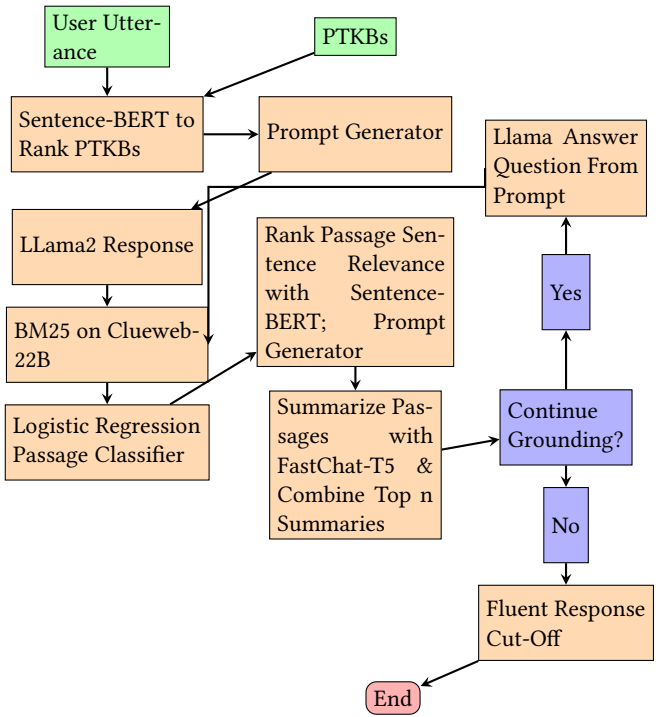
\begin{figure}[t!]
\begin{center} 
    \begin{tikzpicture}[node distance=1.3cm]
\node (userutterance) [io] {\begin{varwidth}{6em}User Utterance\end{varwidth}};
\node (ptkbs) [io,right of=userutterance,xshift=2cm] {\begin{varwidth}{6em}PTKBs\end{varwidth}};
\node (sentance_bert) [process, below of=userutterance] {\begin{varwidth}{8em}Sentence-BERT to Rank PTKBs\end{varwidth}};
\node (query_generator) [process, right of=sentance_bert, xshift=2cm] {\begin{varwidth}{8em}Prompt Generator\end{varwidth}};
\node (llama) [process, left of=query_generator,xshift=-2cm,yshift=-1.5cm] {\begin{varwidth}{8em}LLama2 Response\end{varwidth}};
\node (bm25) [process, below of=llama] {\begin{varwidth}{8em}BM25 on Clueweb-22B\end{varwidth}};
\node (tfidf) [process, below of=bm25] {\begin{varwidth}{8em}Logistic Regression Passage Classifier\end{varwidth}};
\node (relevance) [process, right of=tfidf,xshift=2cm,yshift=2cm] {\begin{varwidth}{8em}Rank Passage Sentence Relevance with Sentence-BERT; Prompt Generator\end{varwidth}};
\node (fastchat) [process, below of=relevance, yshift=-1cm] {\begin{varwidth}{8em}Summarize Passages with FastChat-T5 \& Combine Top n Summaries\end{varwidth}};

\node (twoshot) [decision, right of=relevance,xshift=1.5cm] 
{\begin{varwidth}{8em}Yes\end{varwidth}};

\node (Llama_Answer_Question_From_Prompt) [process, above of=twoshot,yshift=.35cm] 
{\begin{varwidth}{8em}Llama Answer Question From Prompt\end{varwidth}};

\node (is_twoshot) [decision, below of=twoshot,yshift=-.5cm] 
{\begin{varwidth}{5em}Continue Grounding?\end{varwidth}};

\node (oneshot) [decision, below of=is_twoshot] 
{\begin{varwidth}{8em}No\end{varwidth}};

\node (final) [process, below of=oneshot] 
{\begin{varwidth}{8em}Fluent Response Cut-Off\end{varwidth}};
\node (end) [startstop, left of=final,xshift=-1cm,yshift=-1cm] 
{\begin{varwidth}{15em}End\end{varwidth}};


\draw [arrow] (userutterance) -- (sentance_bert);
\draw [arrow] (ptkbs) -- (sentance_bert);

\draw [arrow] (sentance_bert) -- (query_generator);
\draw [arrow] (query_generator) -- (llama);
\draw [arrow] (llama) -- (bm25);
\draw [arrow] (bm25) -- (tfidf);
\draw [arrow] (tfidf) -- (relevance);
\draw [arrow] (relevance) -- (fastchat); 

\draw [arrow] (fastchat) -- (is_twoshot); 
\draw [arrow] (is_twoshot) -- (twoshot); 
\draw [arrow] (twoshot) -- (Llama_Answer_Question_From_Prompt); 
\draw [arrow] (is_twoshot) -- (oneshot); 
\draw [arrow] (oneshot) -- (final); 
\draw [arrow] (final) -- (end); 
\draw[arrow, behind path] (Llama_Answer_Question_From_Prompt.west) -- ++(0,-.25) -| (bm25.east);

\end{tikzpicture}
\end{center} \caption{System Architecture.} \label{fig:architecture} 
\end{figure}

\subsection{Overall Architecture }
\label{sec:architecture}


In this paper, we propose a hybrid method, that combines information retrieval and machine learning methods with state-of-the-art advancements in Language Models across several different architectures. Our solution augmented passage retrieval methods such as BM25 with LLM-enabled response generation and grounding.

We used multiple off-the-shelf models and algorithms, fine-tuning and enhancing some, and creating some of our own models using existing approaches. Llama is the main LLM that we use in this work. We relied on Llama for all conversational aspects of our solution, and for the initial step of answering the question which we could then use for grounding. Because No Llama response was ever used as a final output, we used the less advanced 13B-chat model for performance reasons.


Retrieval is used two times in our architecture. For the ranking of the PTKBs, common approaches include using keyword similarity \cite{ionescu2019vector}, where keywords of the statement are compared against the keywords of the question. This has many drawbacks, including the fact that it cannot handle follow-up questions (for instance, if the user asks "What are the best diets", and the response is a list of 3 diets, if the user asks "What's so great about the 3rd one?", there will be no PTKB matches for statements regarding dietary restrictions). Finally, for the task of retrieving the passages from the corpus, many common solutions place too much reliance on the quality and authenticity of the passages in the corpus. Since most corpora today are scrapped from the internet, they contain data from many malicious or untrustworthy sources. Some solutions will just retrieve passages and use the top N passages in generating their response. This can lead to incorrect or bad-quality responses. 

To improve our results, we also trained a Logistic Regression text classification model using TF-IDF Vectorization, using several datasets representing reliable passages (Wikipedia Articles) and unreliable vs. malicious passages (Retrieved from ClueWeb-22B via Keywords often associated with spam and BM25). 


Figure \ref{fig:architecture} illustrates the architecture of our system. It contains a few components and works in the following steps. (1) Our system first ranks PTKBs with sentence transformers (Sentence-BERT) and combines them together to create a linguistically fluent prompt. (2) We then generated a response from that prompt and the conversation history with Llama, and (3) used that response with BM25 to find relevant passages. In order to keep our LLama's conversation history parallel to the history (the 'canonical responses') found in the test conversations, we used a 'fresh start' approach where the LLM's history was set to be all the previous given conversations as found in the test file, with only the most recent question including our generated prompt. (4) Next, we weeded out bad quality passages with our Logistic Regression Text Quality Filter. (5) We then optimized the top 5 passages by ranking the sentences in order of relevance to write another prompt, again using Sentence-BERT. (6) From there, we took the top 512 characters in order to not overload FastChatT5's \cite{zheng2023judging} context, and summarized each passage in 1-3 sentences. In order to prevent FastChat from drawing from its knowledge base and introducing new information not found in our passages, we kept it blind to the user utterance and gave it a low temperature. We combined the passage summaries and then (7) prompted Llama to generate a response to the user utterance based on the summaries, took that response, and retrieved passages with BM25. 


Note that the retrieval-generate loop can be repeated $X$ times. In our experiments, we tested $X=2$ or $X=1$. When $X=2$, it is to increase the likelihood that Llama would generate something in response to the user utterance that we could find supporting passages for. If Llama asked a follow-up question, then we wouldn't be able to find many passages from its generated response that were actually relevant to the user utterance. In runs with more than 1 shot, we would feed passages into Llama and ask it to answer the utterance from the information contained in the passage. Only on the last cycle would we use the final FastChat summary as the final output.

\subsection {Components}

\subsubsection{Ranking PTKBs by Sentence-BERT} Our system first ranks PTKBs with Sentence-BERT~\cite{reimers2019sentencebert} and combines them together to create a linguistically fluent prompt. Sentence-BERT is ``a modification of the pre-trained BERT network that uses siamese and triplet network structures to derive semantically meaningful sentence embeddings that can be compared using cosine-similarity" \cite{reimers2019sentencebert}. BERT (Bidirectional Encoder Representations from Transformers) models are widely used for deep language modeling and language understanding. BERT can be used in a variety of domains, such as question answering and language inference~\cite{devlin2019bert}. BERT achieves high scores on several Natural Language Processing (NLP) benchmarks, and has the ability to deeply understand language meaning based on context. BERT comes in pre-trained versions, trained on English language text datasets, but can be entirely retrained on other datasets to improve performance on specific domains~\cite{9920081}. BERT also has the advantage of being easily fine-tuned from specialized datasets to fit certain tasks, such as aviation \cite{wang2023adapting}. 



For PTKB ranking and passage optimization, we choose paraphrase-MiniLM-L6-v2~\cite{reimers-2019-sentence-bert} that could both take in up to 512 tokens, needed for passage optimization and because it could deeply consider meanings behind the PTKBs with its 384 feature dense vector representation. 

\subsubsection{Prompting LLMs for Initial Answers}

We then use the prompt created by the ranked PTKBs and the conversation history to prompt an LLM. After extensive testing of many post-GPT 3.5 LLMs, we found two that would work best in specific scenarios: LLaMa 13B-Chat \cite{touvron2023llama} for its chat capabilities and extensive knowledge coverage, and FastChat T5 for passage summarization due to its ability to both take in and accurately understand a larger context and avoid prompt repetition. LLaMa 13B-Chat is used here for prompting for initial answers, using a question-like prompt formed by combining all relevant PTKBs above a threshold with the user utterance in a fluent manner. 



We experimented with using quantized versions of each model, both to allow us to use LLaMa models with more parameters, and to reduce run times.

\subsubsection{Grounding LLM Answers via BM25 Passage Retrieval}

BM25 is a widely used method for retrieving text from search queries. Although more complicated methods exist, such as Learned Sparse and Learned Dense Retrieval Models, BM25 still sometimes ranks above these methods in terms of retrieval accuracy \cite{rosa2021yes}. BM25 builds on top of TF-IDF (Term Frequency-Inverse Document Frequency), which works by finding passages where frequently uncommon terms (terms not contained in most passages) are common \cite{tf-idf}. 
We used the  Lucene \cite{Lin_etal_SIGIR2021_Pyserini} to both build indexes of the iKAT corpus (ClueWeb 22) as well as search in the corpus using BM25. LLM Answers were fed straight into BM25, and by picking the top several passages, we found that our results covered the topics mentioned by the LLM.


\subsubsection{Logistic Regression (LR) for Text Quality Classification}



In order to ensure passage quality, we trained a Logistic Regression model on TF-IDF \cite{tf-idf} vectorizations of text data. Although more advanced methods of text classification exist, Logistic Regression (LR) methods can outperform them on specific tasks. It works by taking labeled training passages and using them to generate vectorizations of each label, then modeling those vectorizations with Logistical Regression to allow for the labeling of new text \cite{tf-idf}.  Logistical Regression is the most thoroughly tested modeling approach for text classification, which under performs on specific tasks but still scores higher than other approaches on most tasks. In the fake-news classifier comparison \cite{vijayaraghavan2020fake}, the researchers tested several approaches for modeling TF-IDF vectors to predict labels, including SVM (Support Vector Machine), Random Forest, LSTM (Long Short Term Memory networks), ANN (Artificial Neural Network), and Logistic Regression. In their tests,  Logistic Regression performed the best, by a margin of ~2\% higher accuracy than the second-best approach. 

To create training data for LR, we labeled an article `reliable' (1000 Wikipedia entries from the WikiText-103 \cite{wikitext}, `unreliable' (good faith passages selected from a list of keywords using BM25 and pyserini \cite{Lin_etal_SIGIR2021_Pyserini} from unreliable sources such as Reddit, Quora, or other opinionated or informal sources), and  `junk' (malicious passages that were written like advertisements, related to spammy topics, used keyword stuffing, or otherwise matched with keywords frequently associated with bad sources). We threw out any passage labeled by our model as either `unreliable' or `junk', keeping only Wikipedia-like passages. Wikipedia-like sources are rather reliable and very useful for QA purposes \cite{chen2017reading}. With a constrained corpus, with a much smaller subset of passages than that of the full collection, we found this worked well for selecting passages that were literally from Wikipedia, or generally written in an encyclopedic style. However, with the full corpus, we found that the top 5 BM25 passages would almost always be from reputable sources, testifying to the proven quality of Clueweb-22B \cite{overwijk2022clueweb22}.


%

\subsubsection{Prompting Again with T5 for Answer Summarization}

Large Language Models have enhanced capabilities for summarizing large passages of text, and thus are commonly used for Question Answering (QA) and other tasks related to information retrieval. Although models like ChatGPT can be the most frequently used model for summarization, newer Text-To-Text models have a much broader set of capabilities, and have ranked highly on summarization tests. Text-To-Text Transfer (T5) based models work by applying what they've learned from tasks to solve related tasks \cite{raffel2020exploring}.  T5 models have several advantages in terms of summarization. T5 Models have the ability to avoid prompt repetition,  since they have a much deeper understanding of textual meaning, and don't rely as heavily on mere text completion. Since they largely aren't used for generating text, and aren't fine-tuned specifically to generate text, they are much more likely to simply reword or shorten given text rather than just generating something new (therefore unreliable) loosely based on the inputted text. Also, T5 models can be adapted to understand much larger contexts than models such as ChatGPT. \cite{xiong2022adapting}. 

After we had weeded out bad quality passages with our Logistic Regression Text Quality Filter, we then optimized the top 5 passages by ranking the sentences in order of relevance to write another prompt, again using Sentence-BERT. Sentence-Ranking has been used before \cite{zhang2020unsupervised} to increase the quality and relevance of text extraction. From there, we took the top 512 characters in order to not overload FastChatT5's \cite{zheng2023judging} context, and summarized each passage in 1-3 sentences. In order to prevent FastChat from drawing from its knowledge base and introducing new information not found in our passages, we kept it blind to the user utterance and gave it a low temperature. 

We tested many post-GPT 3.5 LLMs, we found FastChat T5 works best for passage summarization due to its ability to both take in and accurately understand a larger context and avoid prompt repetition.~\cite{raffel2020exploring} FastChat T5 is also a Text-to-Text Transfer Transformer type model, which is highly customizable and applicable to many applications, including summarization.  We opted to keep FastChat running at full detail, as we would use its summaries for our final product and (in two-shot runs) for our second round of passage selection. FastChat would be given a prompt telling it to summarize the given passage in 1-2 sentences. We found that only very rarely will it go meta and mention the fact that it was summarizing a passage (15/322 times), ideal for our use case.

Looping ensures that the LLM is in alignment with the passages found by BM25. For instance, if the LLM lists five good cars for new parents, BM25 may find passages that mention other cars, in addition to one or two of the cars found by the LLM. LLM-BM25 Congruence helps work to fact-check the LLM and decrease the likelihood of hallucination.


\section{Experiments}

\subsection{Submitted Runs}

\begin{figure}[t]
\raggedright
\begin{sloppypar} 
\caption{Example Turn of Two-Shot Method.} \label{fig:example-full-run}
\texttt{\textbf{User Utterance: } What is a good diet? \\
\textbf{Our Resolved User Utterance: } I'm vegetarian and lactose intolerant, and would like to lose weight. What is a good diet? \\
\textbf{LLaMa Response \#1: } The best diets for a lactose-intolerant vegetarian who would like to lose weight is the ... \\
\textbf{Combined Passage Summaries \#1: } There are many different grains, beans and legumes, grains, vegetables, and plant oils that can be included in a vegetarian diet... \\
\textbf{LLaMa Response \#2: } One such diet that involves beans and legumes, grains, vegetables, and plant oils is the Mediterranean Diet...  \\
\textbf{Final Answer/Combined Passage Summaries \#2: } The Mediterranean diet is a plant-based eating plan that includes lots of fruits, vegetables, beans, lentils, nuts, and whole grains... \\ 
}
\end{sloppypar} 
\end{figure}

We submitted three runs for TREC 2023 evaluation. Our first run consisted of a two-shot approach in which we completed two cycles of the steps described in Section \ref{sec:architecture}. 

Our second run used only one shot, and therefore just generated one combined passage summary. 

Our third run was the same as the second run, in that it was a one-shot approach, with the slight difference that we did not use the LR to select text quality, therefore relying soley on the passages ranked by BM25. We initially tried to use an LLM to certify passage relevance, both by inputting the full passage as well as the passage summary into the LLM, asking about its relevance to the user utterance. We opted to not use this step in our final solution, as all models we tried had a very high error rate for identifying whether a passage was relevant or not.

\subsection{TREC Evaluation}

\begin{table}[h]
\centering
\caption{Scores for All Topics for our runs and the Averaged Medians for All Runs.}   \label{fig:results} 
  \begin{tabularx}{\linewidth}{
    |>{\hsize=1\hsize}X|
    >{\hsize=1\hsize}X|
    >{\hsize=1\hsize}X|
    >{\hsize=1\hsize}X|
  }
    \hline
      & \textbf{success\_1} & \textbf{ndcg\_cut\_5} & \textbf{ndcg\_cut\_10} \\
    \hline
    Our Run 1 & 0.4229 &  0.2626 & 0.2707 \\
    \hline
    Our Run 2 & 0.5170 &  0.3059 & 0.3154 \\
    \hline
    Our Run 3 & 0.5341 &  0.3233 & 0.3216 \\
    \hline
    Median Averages & 0.2074 &  0.123 & 0.1266 \\
    \hline
\end{tabularx}

\end{table}

Across the three tests for which median scores were made available, \textit{Success 1, nDCG at 5}, and \textit{nDCG at 10}, each of our runs' average scores for all topics beat the averages of the medians of each score across all topics included by a notable margin (Figure \ref{fig:results}). Our best run, by a small margin across all three tests, was the third run, which was the same as the second-ranked run (the one-shot approach), with the only difference being that it did not use the LR passage quality classifier. Success\_1 is measured as either a 0 or a 1, and is likely a measure of whether or not the response met the requirements of a `correct' response. NDCG, or (Normalized Discounted Cumulative Gain) is a measure of ranking quality independent of the particular query, and the cut\_5 or cut\_10 scores represent how many of the top documents were relevant \cite{wang2013theoretical}. Our one-shot methods (Runs 2 \& 3) beat our two-shot method (Run 1), and this could be because one bad quality step could throw off the remaining steps throughout the cycle, resulting in a worse answer. For instance, if Llama failed to answer the question adequately, no good supporting passages would be found, leaving Llama with a nonsense input that would result in a bad, ungroundable output. It could also hint at Llama being inept in generating responses based on given information. Since the only difference between Run 2 and Run 3 is that Run 2 uses the passage quality classifier and Run 3 does not, and since Run 3 scored higher than Run 2, our passage quality classifier did not help in generating better responses. This could be because the rather primitive, bag-of-words-based methods we used to judge passages were incapable of adapting to the type of passages found in ClueWeb-22B. Additionally, it could be because this passage quality layer was not needed, and that BM25 was able to adequately find good-quality passages on its own. 

\subsection{Limitations of Existing Solutions}


Throughout the research and development of our solution, we discovered several caveats to popular models, methods, and algorithms. For instance, if any LLaMa model is given more than 2048 tokens it will not be able to coherently respond and will often output gibberish (Table \ref{table:llama-gibberish}). With FastChat, we found that quantization had a large impact on the quality of outputs (Table\ref{table:fastchat-quantization}), and that ordering sentences based on relevance to user utterance greatly improved passage summaries (Table \ref{table:relavance-optimization}). Quantization has been found to noticeably impact results \cite{zhu2023survey}, but was necessary for preformance reasons. Passages often contain a lot of information that isn't relevant to the user utterance. For instance, we found an article about the best 10 natural diets, and the user was asking about one specific diet that was included in that list, Without sentence ranking the response would not adequately address the user utterance. Sentence ranking allowed us to keep FastChat blind to the user utterance in that it wouldn't try to generate a response based on passage summaries to it like we did with Llama in the first round of our two-shot approach. Using Large Language Model outputs carry a number of risks, such as hallucination, where they generate false statements, and in our use case, would not be fully grounded in our passage collection. We also found that ChatGPT and other such chat-based LLMs were not able to adequately grasp whether or not a given passage was relevant.

\begin{table}[t]
\caption{FastChat-T5 Summary of `Large' Wikipedia Snippets (around 550 Tokens/2600 Characters)) asking about a small detail in the passage.} \label{table:relavance-optimization} 
  \begin{tabularx}{\linewidth}{
    |>{\hsize=1\hsize}X|
    >{\hsize=1\hsize}X|
    >{\hsize=1\hsize}X|
  }
    \hline
    \textbf{Question [notes, not given to LLM]} & \textbf{No Optimization}  & \textbf{With Optimization} \\
    \hline
    Which planets have weather? & The inner Solar System consists of the terrestrial planets and the asteroid belt, which are composed mainly of silicates and metals.& Three of the four inner planets (Venus, Earth, and Mars) have atmospheres substantial enough to generate weather. \\
    \hline
    What was the original planned cost for the Macintosh classic? & The Macintosh Classic was produced by Apple in response to the success of previous Macintosh computers & The original Macintosh plans called for a system around \$1,000 \\
    \hline
    What caused the Boeing 737 Max 8 to crash? & The Boeing 737 MAX was certified by the FAA in 2016 and has been involved in two crashes & The crashes of the Lion Air and Ethiopian Airlines were attributed to faulty aircraft design and other factors, including maintenance and flight crew actions. \\ 
    \hline
  \end{tabularx}
\end{table} 

\begin{table} [t] 
\caption{LLama2 13B-Chat Generating Gibberish} \label{table:llama-gibberish} 
\centering
\begin{sloppypar}
\texttt{
\textbf{A: }Tracebackership Tracebackhips Tracebackteenagersees:\u2009eyampholderstandinggoingways \\ 
}
\end{sloppypar} 
\vspace{.1in}
\end{table} 

\begin{table}[t]
\centering
\caption{Eight-Bit Quantization FastChat-T5 Summaries vs. Full Quality Summaries of `Large' Wikipedia Snippets}  \label{table:fastchat-quantization} 
  \begin{tabularx}{\linewidth}{
    |>{\hsize=1\hsize}X|
    >{\hsize=1\hsize}X|
    >{\hsize=1\hsize}X|
  }
    \hline
    \textbf{Question Type}  & \textbf{Quantized} & \textbf{Full Quality} \\
    \hline
    Generating Passage Summary & John J. DeGioia is the president of Georgetown University, which was founded in 2001 [incorrect] & John J. DeGioia is the president of Georgetown University, which he has led since 2001... [correct] \\
    \hline
\end{tabularx} 
\end{table}  

\begin{table}[t]
\centering
\caption{Results of a Test Retrieve-Then-Generate Run}  \label{table:Retrieve-Then-Generate-Run-Scores} 
  \begin{tabularx}{\linewidth}{
    |>{\hsize=1.15\hsize}X|
    >{\hsize=0.95\hsize}X|
    >{\hsize=0.95\hsize}X|
    >{\hsize=0.95\hsize}X|
  }
    \hline
      & \textbf{success\_1} & \textbf{ndcg\_cut\_5} & \textbf{ndcg\_cut\_10} \\
    \hline
    Retrieve-Then-Generate Run & 0.1477 & 0.0882 & 0.0618 \\
    \hline
    Our Run 3 (our best run) & 0.5341 &  0.3233 & 0.3216 \\
    \hline
    Median Averages & 0.2074 &  0.123 & 0.1266 \\
    \hline
\end{tabularx}
\end{table}  

\subsection{Discussion}

Since TREC iKAT submissions are required to be ranked in order of quality, we used several methods to evaluate a myriad of runs to both select and rank our top three. We took several randomly selected conversations/topics and evaluated each for several factors: Passage Quality, Passage Relevance, Summary Relevance, and Conversationally. We ranked our two-shot solution above our one-shot solution due to its ability to self-correct for a previous shot's misstep. For instance, if LLaMa asked a follow-up question to the user utterance not asked in the canonical responses, then there would be no relevant information to select passages for. The second shot would give Llama another chance to correct itself, with the added information from the less-than-relevant passages pushing Llama to provide more information. Additionally, we decided to submit a run without using the LR based passage quality classifier as, due to the very basic nature of the TF-IDF term weighting, its accuracy may have been limited when dealing with passages in the gray area between relevant and junk/untrustable. As shown below, our self-evaluated order of runs turned out to be the exact opposite of how TREC rated each of them. This could be because we ordered our runs in terms of complexity, in other words, always assuming runs with more steps/layers would be better than simpler runs. 

In order to determine how much impact our Generate-Retrieve-Generate approach had on our scores, we conducted an almost identical run, with the only difference being the order of our retrieval and generation. We would take our self-resolved user utterance and find passages with BM25, then generate summaries from those passages. As shown in Table \ref{table:Retrieve-Then-Generate-Run-Scores}, our Generate-Retrieve-Generate Approach greatly outpaced a test run that used a Retrieve-Then-Generate method. Our best run scored 280\% higher than this run in the Success 1 test, 300\% higher in nDCG at 5, and 433\% in nDCG at cut 10. The average median scores for all submitted runs were also higher than the scores of this test run, with the median having a 40\% higher Success 1 score, a 50\% higher score for nDCG at cut 5, and a 100\% higher score for nDCG at cut 10. Median scores outpacing scores for this test could suggest that those using Retrieve-Then-Generate-Run-Scores could have been using more advanced or superior components for parts of their design than we were.

\section{Conclusion}

In conclusion, our solution to TREC-iKAT relied on human-generated content for factual responses to avoid LLM hallucination and bias, whilst still being highly conversational via a hard reliance on LLaMa 2 13B-Chat, a post-GPT 3.5 open-sourced LLM. We used a diverse collection of several different Language Models, including BERT-based sentence transformers for text classification and understanding, and a Text-to-Text Transfer-based model for passage summarization. We used a Generate-Retrieve-Generate approach, which allowed us to combine the best aspects of traditional retrieval methods with those of generation-based methods. During the development of our solution, we encountered several quirks and drawbacks associated with common models and approaches, like Llama's inability to summarize and how much sentence optimization helps improve FastChat summarization approaches. We also realized the importance of using several different LLM architectures for multi-faceted tasks. General purpose models, like ChatGPT, tended to under preform purpose-built models for hyper-specific tasks like sentence-similarity ranking and full-truth summerization. Additionally, we saw that traditional methods like BM25 can still out preform much more advanced and modern methods.  

Although our passage filtering and summary generation methods were rather complex, our reliance on standard and rather basic algorithms such as BM25 for our passage retrieval and logistic for our text classifier, left many areas for improvement. The scope of our solution was limited by several factors, such as our hardware and time. With better hardware, we could have used full-quality models of LLaMa and FastChat-T5 without having unreasonable runtimes and could have generated more runs and evaluated more potential improvements within the time constraints of the competition. 

\section*{Acknowledgement}

This research was supported by the Dehejia Fellows Internship Program. Any
opinions, findings, conclusions, or recommendations expressed in this paper are of the authors and do not necessarily reflect those of the sponsor.

\printbibliography

\end{document}